\documentclass[a4paper,conference]{IEEEtran}
\usepackage{cite}

\ifCLASSINFOpdf
   \usepackage[pdftex]{graphicx}
   \graphicspath{{./graphics/}}
   \DeclareGraphicsExtensions{.pdf,.jpeg,.png}
\else
   \usepackage[dvips]{graphicx}
   \graphicspath{{./eps/}}
   \DeclareGraphicsExtensions{.eps}
\fi
\usepackage{amsmath}
\usepackage{algorithmic}

\usepackage{array}

\ifCLASSOPTIONcompsoc
  \usepackage[caption=false,font=normalsize,labelfont=sf,textfont=sf]{subfig}
\else
  \usepackage[caption=false,font=footnotesize]{subfig}
\fi
\usepackage{fixltx2e}

\usepackage{stfloats}
\usepackage{url}

\usepackage{booktabs} 
\usepackage{tabularx} 
\usepackage{multirow, makecell} 
\usepackage{caption}
\usepackage{csquotes} 
\usepackage{dsfont}

\begin{document}
%
\title{Shuffle \& Divide: \\
	Contrastive Learning for Long Text}

\author{\IEEEauthorblockN{Joonseok Lee, Seongho Joe, Kyoungwon Park, Bogun Kim, Hoyoung Kang, Jaeseon Park, Youngjune Gwon}
\IEEEauthorblockA{Samsung SDS, Seoul, South Korea  \\
	Email: \{js1985.lee, drizzle.cho, kw621.park, bogun0.kim, hoyoung.kang, jaeseon.park, gyj.gwon\}@samsung.com}}


%


\maketitle

\begin{abstract}
We propose a self-supervised learning method for long text documents based on contrastive learning. A key to our method is \emph{Shuffle and Divide} (SaD), a simple text augmentation algorithm that sets up a pretext task required for contrastive updates to BERT-based document embedding. SaD splits a document into two sub-documents containing randomly shuffled words in the entire documents. The sub-documents are considered positive examples, leaving all other documents in the corpus as negatives. After SaD, we repeat the contrastive update and clustering phases until convergence. It is naturally a time-consuming, cumbersome task to label text documents, and our method can help alleviate human efforts, which are most expensive resources in AI. We have empirically evaluated our method by performing unsupervised text classification on the 20 Newsgroups, Reuters-21578, BBC, and BBCSport datasets. In particular, our method pushes the current state-of-the-art, SS-SB-MT, on 20 Newsgroups by 20.94\% in accuracy. We also achieve the state-of-the-art performance on Reuters-21578 and exceptionally-high accuracy performances (over 95\%) for unsupervised classification on the BBC and BBCSport datasets.
\end{abstract}


 \ifCLASSOPTIONpeerreview
 \begin{center} \bfseries EDICS Category: 3-BBND \end{center}
 \fi
%
\IEEEpeerreviewmaketitle

\section{Introduction}

Self-supervised Learning (SSL) can provide label-equivalent information necessary in gradient descent updates to a neural network. A carefully-designed pretext task facilitates the source of label-equivalent information. In a modern paradigm for natural language processing (NLP), an upstream task is set up in a self-supervised manner to pre-train a model, followed by a downstream task that is fine-tuned on the pre-trained model. The upstream task typically requires no labeled data examples (large text corpora) to build a language model, whereas the downstream task can be fine-tuned on a relatively smaller number of labeled examples. 

In NLP, Transformer~\cite{vaswani2017attention} has enabled some important SSL methods that produce outstanding results. The Transformer-based models such as BERT~\cite{Devlin2019BERTPO} and ALBERT~\cite{lan2019albert} have empirically demonstrated the effectiveness of SSL. Up until now, however, its success is mainly for upstream tasks, not for the downstream. It requires yet another labeled data to fine-tune the pre-trained models. In other words, it still depends on supervised learning that requires much manual effort. NLP downstream tasks and their performances are limited by disadvantages of supervised learning.

In image processing, on the other hand, numerous approaches to utilize SSL have been fruitful. Especially, contrastive learning method has shown great advances in image classification with massive unlabeled data. It aims to use similarities and differences between images. SimCLR~\cite{chen2020simple} is a prominent work for this approach. As its pre-trained model applies to downstream task of classification, it outperforms even supervised learning models. But, it also need two stage process of upstream and downstream task.

To take one step further, some works make downstream task also self-supervised, to minimize human intervention. They include SCAN~\cite{van2020scan}, RUC~\cite{park2021improving}, SelfMatch~\cite{kim2021selfmatch}, yielding tangible results.

There is not much work to extend SSL or unsupervised learning method to downstream task in natural language processing field. A few has tried contrastive learning in the field. Although image can be augmented quite naturally and many image augmentation techniques already exist, natural language has relatively little techniques. Naturally, many has tried SSL with sentence similarity. Basic way of it is to use TF-IDF of documents. But it simply regards documents with the same frequency of a token similar. Moreover, it cannot distinguish homonym.

To overcome these shortcomings, some approaches have tried with semantic sentence similarity. They are G-BAT~\cite{wang2020neural} and SS-SB-MT~\cite{Chiu2020AutoencodingKC}. SS-SB-MT yields the state-of-the-art performance in unsupervised text classification. It generates keyword correlation graph with edges and nodes by using sentence similarity (SS) and Sentence BERT (SB). Then, multi-task graph autoencoder (MT) transform the graph into latent feature, which is to get document clusters. In short, it converts document features into graphs, then gathers similar ones together.

Pioneering work of SS-SB-MT outperforms previous works in long sentence classification with SSL. However, it only learns graphs summarizing documents, not the whole text. It is not effective for exploiting information of documents, compared to methods to learn feature vectors from whole text. New way of generating supervision effectively by conserving semantic features is needed.

We propose a new contrastive learning technique for document classification and corresponding effective sentence augmentation in this paper. Contrastive learning is to build positive relation between input data and its augmented ones, and negative with the others. So, optimal way of data augmentation is crucial. One method is back translation, which translates back to original language. But translation could be noisy. Moreover, its result is very similar to the input, i.e., weak augmentation.

To overcome previous limitations, we have developed new augmentation for contrastive learning: shuffling whole sentences in a document, then dividing them into two groups. It conserves meaning of individual sentence, simultaneously augments input strongly. Strong augmentation with whole semantics can make contrastive training optimal.

Experiments with 20 news group datasets for document classification task shows classification accuracy 68.34~\%, outperforming G-BAT (41.30~\%) and SS-SB-MT (47.40\%) by 27.04~\%p and 20.94~\%p improved respectively, achieving new state-of-the-art. Experiments with datasets like router also shows outstanding results.

In conclusion, we suggests novel SSL techniques for unsupervised document classification, resulting in state-of-the-art accuracy performances. We have empirically demonstrated the efficacy of contrastive learning in unsupervised text classification.

In the paper, our contributions are summarized as follows.

\begin{itemize}
	\item We present state-of-the-art contrastive SSL for long text (intent) classification;
	\item We suggest contrastive learning can apply to tasks in natural language processing;
 	\item We propose Shuffle and Divide (SaD), a novel text augmentation algorithm for effective contrastive learning of document classification.
\end{itemize}

The rest of this paper is organized as follows. Section 2 describes related work. In Section 3, we explain our architecture and algorithms in detail. Section 4 presents our experimental results and gives quantitative comparison with previous work. The paper concludes in Section 5.

\section{Related Work}

We review widely-used approaches, self-supervised learning and text augmentation, which are related to our work in this session.

\subsection{Self-supervised learning}

Self-supervised learning is to learn representation with unlabeled large-scale datasets. Traditionally, most deep learning models have relied on supervised learning. But they have required human supervision. SSL techniques includes MLM(Masked Language Model)~\cite{Devlin2019BERTPO}, NSP(Next Sentence Prediction)~\cite{Devlin2019BERTPO}, and contrastive learning~\cite{chen2020simple}. They uses tasks which can be defined with unlabeled data, i.e, pretext task. A model pre-trained with pretext task is transferred to domain-specific situations. It is called downstream task.

Pre-training methods for SSL includes context prediction~\cite{doersch2015unsupervised}, image colorization~\cite{zhang2016colorful}, Jigsaw puzzle solving~\cite{noroozi2016unsupervised}, and rotation prediction~\cite{gidaris2018unsupervised}. As contrastive learning based on InfoMax principle~\cite{tschannen2020mutual}, SSL for image have advanced, with the works of CPC~\cite{oord2018representation}, DIM~\cite{hjelm2019learning}, AM-DIM~\cite{bachman2019learning}, MoCo~\cite{he2020momentum}, and SimCLR~\cite{chen2020simple}. Pre-trained models for text SSL includes BERT~\cite{Devlin2019BERTPO} and GPT-3~\cite{brown2020language}

\paragraph{BERT}

A language representation model BERT uses only encoder part of transformer~\cite{vaswani2017attention}. Language pre-trained models have outperformed previous ones in various natural language tasks~\cite{Dai2015SemisupervisedSL}. With unlabeled large-scale dataset, BERT defines two pretext tasks: MLM and next sentence prediction. MLM is to predict randomly masked words by learning context of sentences or words with attention. Next sentence prediction is to infer relationship between given two sentences.

\paragraph{SimCLR}

SimCLR learns visual representation from unlabeled large-scale datasets with SSL method. It applies two different augmentations for each image. Then, augmentations from the same image becomes positive pair, the others negative pair, contrasting them to learn features. Its performance come close to supervised learning models.

\subsection{Text augmentation}

Data augmentation aims to generate new data from the original, conserving their labels. For text, it includes text modification and generation. Examples of modification are random noise injection~\cite{wei2019eda}, randomly removing, inserting, or replacing words, and TF-IDF based word replacement, modifying less important words which are not in principal component of TF-IDF matrix. Examples of text translation are Back translation~\cite{Edunov2018UnderstandingBA}, translating back and forth, and generative methods~\cite{kumar2020data}, using sentence-generating models.

\subsection{Document clustering}

Embedding for documents should be extracted, to cluster them. Traditional ways of embedding are well-known, such as TF-IDF~\cite{sammut2010tf}, which counts the occurrence of words per document to give weight, LSI (Latent Semantic Indexing)~\cite{papadimitriou1998latent}, LDA (Latent Dirichlet Allocation)~\cite{blei2003latent}, and glove embedding average. Deep learning approaches have added new options. BAT, G-BAT~\cite{wang2020neural} and SS-SB-MT~\cite{Chiu2020AutoencodingKC} use neural networks to embed documents, to outperform traditional clustering methods. According to \cite{Chiu2020AutoencodingKC}, ELMo~\cite{peters2018deep} and SBERT~\cite{reimers2019sentence} extract embedding of sentences or words, then averaging them to get better embedding.

\paragraph{BAT and G-BAT}
BAT trains generative adversarial network (GAN) with TF-IDF vector and topic distribution. The encoder of that network is re-used to get embedding of documents to cluster. G-BAT, which apply multivariate Gaussian to the generator model of BAT, outperformed traditional methods on 20NG datasets.

\paragraph{SS-SB-MT}
SS-SB-MT is to build keyword correlation graphs with sentence similarity (SS) and sentence BERT (SB). Multi-task graph autoencoder (MT) extracts latent features form graphs. Document clustering works in that feature space. It is innovative, but quite complicated model to handle. It achieved previous state-of-the-art on 20 newsgroup and Reuters datasets.

\section{Method}

Since contrastive learning does not depend on the model architecture, it can be used to enhance a pre-trained NLP model in a specific way by running large corpora, all without additional labeled data. In NLP, however, it is difficult to make positive and negative examples necessary for contrastive learning by using augmentation, unlike SimCLR used for vision applications.

By overcoming the limitations of text augmentation, we propose a method of fine-tuning the NLP model in a contrastive self-supervised method without additional labels.

\subsection{Model Architecture}

\begin{figure}[h]
    \begin{center}
    \includegraphics[width=\linewidth]{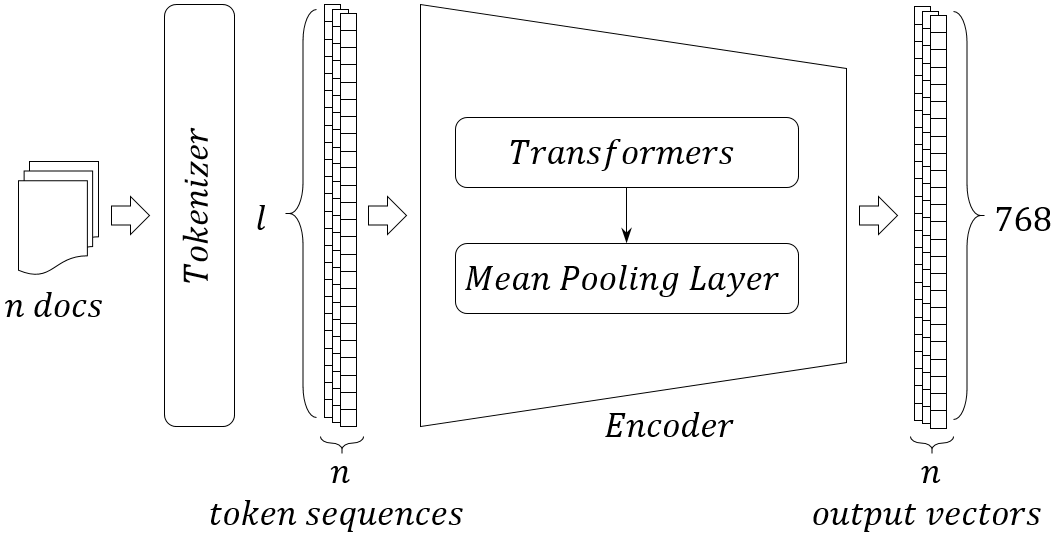}
    \end{center}
    \caption{Model Architecture. The encoder model receives $n$ documents and outputs 768-dimensional vectors. $l$ is the max sequence length and is predefined. Encoder consists of transformers and Mean Pooling Layer. Various models such as BERT and RoBERTa can be used as Transformers. The mean pooling layer calculates the mean of all output vectors like that of Sentence BERT.}
    \label{fig:architecture}
\end{figure}

\begin{figure*}[t]
	\begin{center}
		\includegraphics[width=\textwidth]{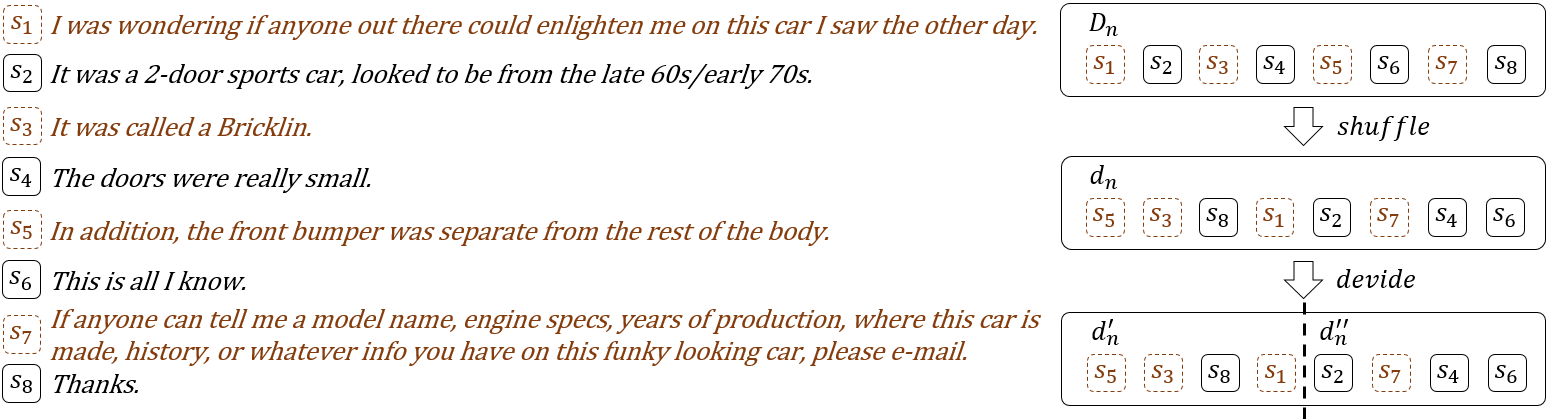}
	\end{center}
	\caption{Shuffle and Divide process for sample document $D_{n}$. Suppose we have a document $D_{n}$ with $m$ sentences from $s_{1}$ to $s_{m}$. First, tokenize $D_{n}$ in sentence units. Then, randomly shuffle the sentence tokens and combine them again to make $d_{n}$. In this case, the meaning or category of the document is not significantly damaged. Next, $d_{n}$ is divided in half to make $d'_{n}$ and $d''_{n}$. If the number of sentences is sufficient, the meaning or category of document $d'_{n}$ and document $d''_{n}$ will also coincide.}
	\label{fig:shuffle_divide}
\end{figure*}

Figure~\ref{fig:architecture} shows the overall architecture of our model. Transformers are initialized with pre-trained weights from the BERT and RoBERTa models. To obtain the embedding of documents required for clustering, we add a mean pooling layer to the Transformers model and use it as an encoder. Mean pooling layer calculates the average of the Transformer output vectors. As a result, the encoder receives a token sequence of a predefined length as input and outputs a 768-dimensional vector.

\subsection{Contrastive Self-supervised Learning}

We use a contrastive self-supervised learning approach to learn latent representations of documents without labels to ultimately cluster them.

During the training process, within each mini-batch, the distance between embeddings of the positive pair is made close to each other, and the distance between embeddings of the negative pair is made farther from each other. Among all document pairs that can be combined in each mini-batch, all pairs that are not positive pairs become negative pairs. Therefore, the same number of positive pairs as the batch size and (batch size * 2 - 2) negative pairs are created for each mini-batch. Then, the token sequences of all documents in the mini-batch are fed to the encoder to obtain output vectors, and the contrastive loss is calculated based on the previously obtained positive and negative pair configurations. The model parameters are updated through the loss calculated in each mini-batch.

In a general contrastive self-supervised method, positive pairs are created through augmentation. In Computer Vision, there are various augmentation options such as rotation, flipping, and color jittering, and by combining them, numerous label-preserving augmented versions of one sample can be obtained. However, In NLP, simple augmentation tricks have little effect in contrastive learning. In order to make meaningful changes to the original sample, it is necessary to use a word dictionary or a trained model such as WordNet Augmenter, Contextual Augmenter, or back translation. However, the effect is limited because the content and sentence structure of the original document and the augmentation result are very similar.

We propose methods suitable for use in contrastive self-supervised learning to replace or improve these existing augmentation methods. The first method is to completely replace text data augmentation. Based on the TF-IDF feature of documents, the pair with the highest cosine similarity is sampled to create a positive pair. Our main approach, the second method, called Shuffle \& Divide, simply shuffles the order of the sentences in the documents and splits the document in half to get a positive pair by splitting one document into two.

\subsection{Positive Sampling with TF-IDF}

The first method to solve the limitations of text augmentation is to completely replace augmentation itself with positive sampling using TF-IDF. Even if augmentation is not used, a contrastive learning method can be applied if a positive pair can be created from the dataset.

Instead of text augmentation, we propose a method to perform contrastive learning by making two similar samples as a positive pair. TF-IDF is a method of assigning weights to each word by using the number of documents in which a specific word appears and the number of times in which a specific word appears in a specific document. It is a traditional method often used to compare similarities between documents.

We will create a TF-IDF vector for all documents in the dataset and define it as set $\{D_n\}$. For all $D_n$, find $D_m(m \neq n)$ with the highest cosine similarity and assume $D_m$ and $D_n$ as a positive pair. Since the positive pair created in this way has a high probability of belonging to the same category but is completely different from each other, it can solve the problem that the result of text augmentation is very weak.

\subsection{Shuffle \& Divide}

\begin{figure}[h]
    \begin{center}
    \includegraphics[width=\linewidth]{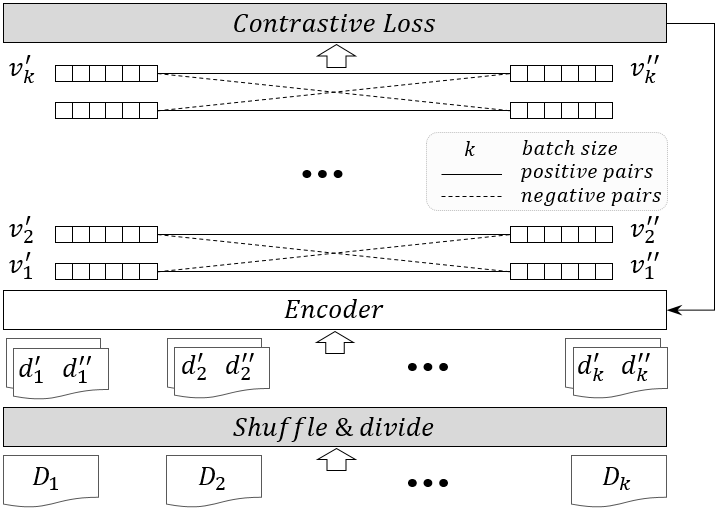}
    \end{center}
    \caption{The process of training the model using the Shuffle \& Divide method. Positive pairs are prepared as many as the number of batch sizes, and negative pairs are prepared by subtracting one positive sample from twice the batch size. In the figure, the negative pair relationship is expressed only for $v_1$ and $v_2$, which are the output vectors of $d_1$ and $d_2$, for visibility. The parameters of the encoder are updated by calculating the contrastive loss in each mini-batch.}
    \label{fig:text_model}
\end{figure}

Positive sampling by TF-IDF completely replaces text augmentation, enabling contrastive self-supervised learning even when augmentation is impossible or insufficient. However, there is a fundamental and potential limitation that positive pairs may actually be of different classes. In fact, when top-1 cosine similarity positive sampling is performed based on the TF-IDF vector of documents in the 20NG dataset, the case where the labels of the positive pair match is about 85\%.

We propose the Shuffle \& Divide algorithm as an alternative to this problem. Figure~\ref{fig:shuffle_divide} is an example of applying Shuffle \& Divide to document $d$. The sentence order of $d$ is shuffled and then divided in half to make $d'$ and $d''$. Since they belong to the same category, they can be used as a positive pair. Shuffle \& Divide is a very simple and intuitive method, but it showed a very powerful effect in the experimental process described in Experiments.

Figure~\ref{fig:text_model} shows the Shuffle \& Divide process. In the representation learning process, only documents consisting of 4 or more sentences are used as training data. Let $D_{n}$ be one of the documents sampled from each mini-batch. When $D_{n}$ is a document composed of $m$ sentences from $s_{1}$ to $s_{m}$, $d_{n}$ is created by randomly shuffled sentence order in $D_{n}$. By dividing $d_{n}$ in half, documents $d'_{n}$ and $d''_{n}$ composed of $n/2$ sentences are created, and a positive pair ($d'_{n}$, $d''_{n}$) can be obtained per sample document $D_{n}$.

In other words, we can get as many positive pairs as the batch size in the mini-batch. Also, since the sentence order of the document is randomly shuffled at every epoch and then divided, the augmentation effect of positive pairs is amplified in the contrastive concept.

\newcolumntype{L}[1]{>{\raggedright\let\newline\\\arraybackslash\hspace{0pt}}m{#1}}
\newcolumntype{R}[1]{>{\raggedleft\let\newline\\\arraybackslash\hspace{0pt}}m{#1}}

\section{Experiments}

To verify that our proposed method works, we have performed clustering on text datasets. Because the output vector from feeding the tokens sequences of documents to the encoder is clustered as is, the performance of the encoder can be observed intuitively. Additionally, for a specific dataset, we compared whether the encoder fine-tuned with our method was better initialized than the ``bert-base-uncased" pre-trained weight of BERT. We prove the effectiveness of our method by adding a classification head to each of the two encoders, performing fine-tuning with the supervised method, and comparing them.

\subsection{Datasets}

\begin{table}[h]
	\begin{center}
		\begin{tabular}{lrrr}
			\toprule
			\noalign{\smallskip}
			\makecell{Datasets} & \makecell{Size} & \makecell{Classes} & \makecell{Length} \\

			\cmidrule(lr){1-1} \cmidrule(lr){2-4}
			\noalign{\smallskip}
			20NG & 18.612 & 20 & 245 \\
			Reuters & 7,316 & 10 & 141 \\
			BBC & 2,225 & 5 & 26 \\
			BBCSport & 737 & 5 & 345 \\
			\bottomrule
		\end{tabular}
	\end{center}

	\caption{Datasets summary.}
	\label{tab:datasets_summary}
\end{table}

We use 20 Newsgroups (20NG)~\cite{Lang1995NewsWeederLT}, Reuters-21578 (Reuters)~\cite{Lewis2004RCV1AN}, BBC and BBCSport datasets for our experiments. To compare with SS-SB-MT~\cite{Chiu2020AutoencodingKC}, which is the current state-of-the-art for unsupervised text classification, 20NG and Reuters are preprocessed according to the method of~\cite{Ye2017DeterminingGA}, same as SS-SB-MT. Preprocessing the 20NG dataset is summarized as follows. The header and footer of each document were removed, and the URL and email address were also removed. And documents with fewer than 10 words were removed. For the Reuters dataset, multi-labeled data was removed, and after removing documents with empty bodies, duplicate data were also removed. And, only the top 10 categories of documents were extracted.

In case of Reuters, the class imbalance is more severe than 20NG, so the randomness in performance measurement tended to be large. The BBC and BBCSport datasets have not been widely used in previous deep learning studies due to the small number of data, but only the Shuffle and Divide method has been tested experimentally. Table~\ref{tab:datasets_summary} summarizes the information of each dataset used in the experiment.

\subsection{Clustering}

In the process of grouping documents expressed as high-dimensional vectors, the definition of distance metric between documents is very important. It is known that cosine distance is more suitable for clustering of high-dimensional vectors than euclidian distance. We used Spherical k-means, an algorithm that is fast and uses cosine distance as a metric, to cluster the 768-dimensional document vectors output from the encoder.

We compared our method with several traditional methods and recent deep learning methods that achieved state-of-the-art. The performance and settings of the models are well summarized in ~\cite{Xie2013IntegratingDC}.

We used Accuracy (ACC), an indicator used in most previous studies, to measure and compare clustering performance. Using the Hungarian algorithm~\cite{Kuhn1955TheHM}, it is possible to efficiently calculate the maximum accuracy value among all true label and cluster mappings. We also measured AMI for comparison with previous studies, and measured and referenced the silhouette score at every epoch to pick the best model without a label. formula~\ref{eq:accuracy} shows how we calculated ACC.

\begin{equation}
	\label{eq:accuracy}
	ACC = \max_n \frac{\sum_{i=1}^{n} \mathds{1}\{l_i=(c_i)\}}{n}
\end{equation}

\subsection{Representation learning}

\begin{table*}[t]
	\begin{center}
		\begin{tabular}{l ccc ccc ccc ccc}
			\toprule
			\noalign{\smallskip}
			{} & \multicolumn{3}{c}{20NG} & \multicolumn{3}{c}{Reuters} & \multicolumn{3}{c}{BBC} & \multicolumn{3}{c}{BBCSport} \\

			\cmidrule(lr){1-1} \cmidrule(lr){2-4} \cmidrule(lr){5-7} \cmidrule(lr){8-10} \cmidrule(lr){11-13}
			\noalign{\smallskip}
			\makecell{Models} & \makecell{ACC} & \makecell{AMI} & \makecell{SS}
			& \makecell{ACC} & \makecell{AMI} & \makecell{SS}
			& \makecell{ACC} & \makecell{AMI} & \makecell{SS}
			& \makecell{ACC} & \makecell{AMI} & \makecell{SS} \\

			\cmidrule(lr){1-1} \cmidrule(lr){2-4} \cmidrule(lr){5-7} \cmidrule(lr){8-10} \cmidrule(lr){11-13}
			\noalign{\smallskip}
			\multicolumn{13}{l}{\textbf{Document Embedding models}} \\
			\noalign{\smallskip}
			
			TFIDF & 33.70  & 41.70  & - & 35.00  & 45.60  & - & - & 37.60  & - & - & 79.90  & - \\
			LSI & 32.30  & 39.80  &  & 42.00  & 40.00  & - & - & 45.40  & - & - & 84.00  & - \\
			LDA & 37.20  & 28.80  &  & 54.90  & 50.30  & - & - & 15.10  & - & - & 61.60  & - \\
			D2C & - & 49.30  &  & - & 53.40  & - & - & 75.90  & - & - & 81.20  & - \\
			G-BAT & 41.30 & - & - & - & - & - & - & - & - & - & - & - \\
			SS-SB-MT & 47.40  & 53.00  & - & 56.30  & 58.40  & - & - & - & - & - & - & - \\

			\cmidrule(lr){1-1} \cmidrule(lr){2-4} \cmidrule(lr){5-7} \cmidrule(lr){8-10} \cmidrule(lr){11-13}
			\noalign{\smallskip}
			\multicolumn{13}{l}{\textbf{Sentence Embedding models}} \\
			\noalign{\smallskip}
			GloVe & 21.70  & 21.00  & - & 38.50  & 37.10  & - & - & - & - & - & - & - \\
			BERT & 41.90  & 40.50  & - & 47.10  & 42.60  & - & - & - & - & - & - & - \\
			SBERT & 44.10  & 45.10  & - & 51.40  & 52.40  & - & - & - & - & - & - & - \\

			\cmidrule(lr){1-1} \cmidrule(lr){2-4} \cmidrule(lr){5-7} \cmidrule(lr){8-10} \cmidrule(lr){11-13}
			\noalign{\smallskip}
			\multicolumn{13}{l}{\textbf{Ours}} \\
			\noalign{\smallskip}
			\textbf{TPS$^\dagger$} & 66.04  & 63.72  & 58.55  & 41.93  & 42.82  & 42.96  & - & - & - & - & - & - \\
			\textbf{SaD$^\ddagger$} & \textbf{68.34} & \textbf{66.65} & \textbf{61.53} & \textbf{58.17} & \textbf{51.82} & \textbf{51.53} & \textbf{95.46} & \textbf{86.41} & \textbf{47.58} & \textbf{98.51} & \textbf{94.77} & \textbf{52.29} \\
			\bottomrule
		\end{tabular}
	\end{center}

	\caption{Clustering performance of our two methods in comparison to various baseline models. ACC represents accuracy and AMI represents adjusted mutual information. SS is a silhouette score, and within the specified training epochs range, we picked the encoder weight at the time of the highest SS. Among the performance indicators of various methods, ACC is the experimental result reported in ~\cite{Xie2013IntegratingDC} and AMI is reported in ~\cite{Ye2017DeterminingGA}. A - sign indicates that it has not been measured or specified in previous studies. (TPS$^\dagger$ is TFIDF positive sampling, and SaD$^\ddagger$ Shuffle \& Divide.)}
	\label{tab:results}
\end{table*}

\begin{table}[h]
	\begin{center}
		\begin{tabular}{lccccc}
			\toprule
			\noalign{\smallskip}
			\multicolumn{2}{c}{Phase} & 20NG & Re$^\sharp$ & BBC & BBCS$^*$ \\

			\cmidrule(lr){1-2} \cmidrule(lr){3-6}
			\noalign{\smallskip}
			\multirow{2}{*}{TPS$^\dagger$} & train & 256 & 256 & - & - \\
			\cmidrule(lr){3-6}
			& test & 256 & 256 & - & - \\

			\cmidrule(lr){1-2} \cmidrule(lr){3-6}
			\noalign{\smallskip}
			\multirow{2}{*}{SaD$^\ddagger$} & train & 128 & 128 & 64 & 64 \\
			\cmidrule(lr){3-6}
			& test & 256 & 256 & 128 & 128 \\
			\bottomrule
		\end{tabular}
	\end{center}

	\caption{Maximum length of the encoder's input token sequence used in each dataset and phase. (Re$^\sharp$ is Reuters, BBCS$^*$ BBCSport.)}
	\label{tab:maximum-length}
\end{table}

\begin{table}[h]
	\begin{center}
		\begin{tabular}{lcc}
			\toprule
			\noalign{\smallskip}
			\makecell{Base encoder} & \makecell{SaD} & \makecell{Accuracy (\%)} \\

			\cmidrule(lr){1-1} \cmidrule(lr){2-2} \cmidrule(lr){3-3}
			\noalign{\smallskip}
			bert-base-uncased & FALSE & 83.54 \\
			bert-base-uncased & TRUE & 85.40 \\
			SaD encoder & TRUE & 87.88 \\
			\bottomrule
		\end{tabular}
	\end{center}

	\caption{Base encoders for initializing.}
	\label{tab:encoder}
\end{table}

With the two methods we proposed, the model learns a representation for each dataset without using any labels at all. First, the encoder's transformers are initialized with ``bert-base-uncased" pre-trained weights. Contrastive loss is calculated from positive and negative pairs sampled in each mini-batch and the parameters of the model are updated. As the contrastive loss, the nt-xent loss introduced in SimCLR is used. For representation learning, the learning rate of 3e-5 was used in the AdamW optimizer. The larger the batch size, the more diverse the negative views for positive pairs. Therefore, 320, the maximum usable size, was used as the batch size. The range of the total training epoch was determined for each dataset and method, and the model with the highest silhouette score within the range was used as the final model.

\paragraph{TF-IDF positive sampling}

The first method, TF-IDF positive sampling, calculates the cosine similarity between the TF-IDF vectors of all documents included in the dataset, and composes the most similar samples as positive pairs. When a positive pair is configured in this way, the probability that the pair belongs to the same category is calculated, 85~\% for 20NG and 89.23~\% for Reuters. In order to change the positive sample according to the progress of the training epoch, after each epoch, the cosine similarity of each of the TF-IDF vectors and the model output vectors was weighted summed to obtain similarity, and then positive sampling was performed again. formula~\ref{eq:similarity} shows the similarity calculation method according to the progress of the learning epoch. 

In the tokenize phase, 256 was used as the maximum token length for 20NG and Reuters, and 128 was used for BBC and BBCSport. Since the number of positive pairs is half compared to the Shuffle \& Divide method, and even that has no augmentation effect, only 2 to 4 epochs were fine-tuned as recommended by the BERT authors. And at every epoch, all the documents of the dataset were fed to the fine-tuned model, and the output vector was clustered to measure the silhouette score, and finally the model with the highest score was kept. 

As a result, as shown in Table~\ref{tab:results}, it was possible to obtain 65.92~\% accuracy and 63.45~\% AMI high performance for the 20NG dataset, surpassing the existing state-of-the-art. However, this method has limitations in that the categories of positive pairs can be different from each other and that the quantitative effect of augmentation cannot be obtained significantly. Therefore, it showed relatively low performance on small datasets such as Reuters.

\begin{equation}
	\label{eq:similarity}
	\begin{aligned}
	\mathcal{S} ={} & \alpha^{epoch-1}\textrm{sim}(v_{tfidf}) \\
	& + (1 - \alpha^{epoch-1})\textrm{sim}(v_{model})
	\end{aligned}
\end{equation}

\paragraph{Shuffle \& Divide}

The authors of BERT recommend training only 2-4 epochs during supervised fine-tuning of the BERT model for downstream tasks. However, we trained our model for (dataset size / batchsize) epochs so that we can maximize the negative views that have been diversified through contrastive learning and maximize the instance level augmentation effect of positive and negative pairs obtained through shuffle \& divide. 

For shuffle \& divide processing, only documents consisting of 4 or more sentences were used as training data. As in the TF-IDF positive sampling method, the best model was picked by measuring the silhouette score. Since the Shuffle\&Divide method divides the document in half, we set the max sequence length to half that of the test phase in the training phase. The max sequence length for each dataset and phase is shown in the table~\ref{tab:maximum-length}.

As a result, we were able to achieve higher performance than TF-IDF positive sampling for all datasets. For the 20NG and Reuters dataset, ACC improvement was about 20~\%p and about 15~\%p compared to the existing state-of-the-art, respectively. In the case of clustering on the BBC and BBCSport datasets, it showed satisfactory performance of more than 95~\% based on Accuracy.

\paragraph{Classification with supervised fine-tuning}

We additionally attach a classification head to the encoder obtained by training with the Shuffle \& Divide method, and then try classification by the supervised fine-tuning method for 20NG dataset. Three experiments were conducted to check whether the BERT weight fine-tuned by the Shuffle \& Divider method is a good initialization point and whether Shuffle \& Divide is effective in the supervised fine-tuning learning process. 

For the first time, an encoder initialized as ``bert-base-uncased" was fine-tuned in a supervised manner without augmentation. In the second method, supervised fine-tuning was performed by adding only Shuffle \& Divider augmentation to the first method. In the last method, the training data was augmented by the Shuffle \& Divide method while supervised fine-tuning the encoder fine-tuned by the Shuffle \& Divider method. 

As a result, the encoder fine-tuned by the Shuffle \& Divide method showed higher performance after supervised fine-tuning for classification than the encoder initialized to ``bert-based-uncased.'' In addition, when using Shuffle \& Divide augmentation for supervised fine-tuning for classification, the performance was further improved. Detailed performance is listed in Table~\ref{tab:encoder}.

\section{Conclusion}

In this paper, we have described a powerful means to document clustering that can lead to state-of-the-art accuracy performances for unsupervised text classification. There is no decisive way to augment natural language text to learn NLP models in a contrastive self-supervised manner. We propose a method that runs on a simple augmentation algorithm called  Shuffle and Divide (SaD), which provides a facility to enhance a Transformer encoder for document embedding. By coupling contrastive learning with clustering, the final document embeddings result in document clusters that are as discriminative as the text classifier fine-tuned on labeled examples. As we have enhanced the encoder consisting of the Transformer and mean pooling layers with our method, clustering with the latent representation of the documents has verified the effectiveness in text classification tasks. Experiments show that our model achieves better performance than many unsupervised approaches including the state-of-the-art results by SS-SB-MT on the 20 Newsgroups and Reuters datasets. We have achieved a high performance with more than 95\% accuracy for unsupervised classification on the BBC and BBCSport datasets. In addition, our SaD can overcome the limitations of text augmentation in a very simple way, and we are expecting SaD to be applicable to various tasks of learning document embedding as well as contrastive learning.

\bibliographystyle{IEEEtran}
\bibliography{IEEEabrv,text}
%
%
%

\end{document}